\documentclass[10pt,twocolumn,letterpaper]{article}

\usepackage[pagenumbers]{cvpr} 

\usepackage{graphicx}
\usepackage{amsmath}
\usepackage{amssymb}
\usepackage{booktabs}

\usepackage{multirow}
\usepackage [english]{babel}
\usepackage [autostyle, english = american]{csquotes}
\MakeOuterQuote{"}
\usepackage{tablefootnote}
\usepackage{dblfloatfix}

\usepackage[pagebackref,breaklinks,colorlinks]{hyperref}

\usepackage[capitalize]{cleveref}
\crefname{section}{Sec.}{Secs.}
\Crefname{section}{Section}{Sections}
\Crefname{table}{Table}{Tables}

\crefname{table}{Tab.}{Tabs.}

\def\cvprPaperID{*****} 
\def\confName{CVPR}
\def\confYear{2022}

\begin{document}

\title{Role of reward shaping in object-goal navigation}

\author{Srirangan Madhavan\\
UC San Diego\\
{\tt\small smadhavan@eng.ucsd.edu}
\and
Anwesan Pal\\
UC San Diego\\
{\tt\small a2pal@eng.ucsd.edu}
\and
Henrik I. Christensen\\
UC San Diego\\
{\tt\small hichristensen@eng.ucsd.edu}
}
\maketitle

\begin{abstract}
   Deep reinforcement learning approaches have been a popular method for visual navigation tasks in the computer vision and robotics community of late. In most cases, the reward function has a binary structure, i.e., a large positive reward is provided when the agent reaches goal state, and a negative step penalty is assigned for every other state in the environment. A sparse signal like this makes the learning process challenging, specially in big environments, where a large number of sequential actions need to be taken to reach the target. We introduce a reward shaping mechanism which gradually adjusts the reward signal based on distance to the goal. Detailed experiments conducted using the AI2-THOR simulation environment demonstrate the efficacy of the proposed approach for object-goal navigation tasks.
\end{abstract}
\label{sec:intro}

\section{Introduction}
\begin{table*}[!b]
    \begin{minipage}{.49\linewidth}
        \flushleft
        \scalebox{0.52}{
        \begin{tabular}{lcc|ccccc|cc}\toprule
        & \multicolumn{4}{c}{$L \geq 1$} && \multicolumn{4}{c}{$L \geq 5$} \\
        \cline{2-5} \cline{7-10}
        \multirow{2}{*}{Models} & \multirow{2}{*}{$r_{bin}$} & $r_{base}$ & \multicolumn{2}{c}{ours} && \multirow{2}{*}{$r_{bin}$} & $r_{base}$ & \multicolumn{2}{c}{ours} \\
        \cline{4-5} \cline{9-10}
        & & \cite{pmlr-v155-pal21a} & $r_{depth}$ & $r_{bbox}$ && & \cite{pmlr-v155-pal21a} & $r_{depth}$ & $r_{area}$\\
        \hline
        GCN \cite{yang2018visual} & $33.1_{(0.8)}$ & $33.3_{(1.4)}$ & $31.7_{(0.7)}$ & $\mathbf{35.3_{(0.5)}}$ & & $25.0_{(1.4)}$ & $23.5_{(1.6)}$ & $\mathbf{26.9_{(1.1)}}$ & $24.6_{(0.8)}$\\
        SAVN \cite{Wortsman_2019_CVPR} & $34.7_{(0.5)}$ & $\mathbf{40.7_{(1.4)}}$ & $32.2_{(0.9)}$ & $39.6_{(0.8)}$ & & $25.8_{(0.8)}$ & $30.0_{(1.4)}$ & $26.8_{1.3}$ & $\mathbf{31.7_{(1.5)}}$\\
        M\_O \cite{pmlr-v155-pal21a} & $58.8_{(1.0)}$ & $64.1_{(0.7)}$ & $\mathbf{66.4_{(0.3)}}$ & $66.3_{(1)}$ & & $40.6_{(0.6)}$ & $46.6_{(1.6)}$ & $50.5_{(0.7)}$ & $\mathbf{51.5_{(1.3)}}$\\
        M\_R \cite{pmlr-v155-pal21a} & $65.5_{(0.6)}$ & $68_{(0.9)}$ & $\mathbf{77.1_{(0.7)}}$ & $69.7_{(0.9)}$ & & $52.3_{(0.8)}$ & $52.3_{(0.5)}$ & $\mathbf{69.2_{(0.8)}}$ & $57.3_{(1.3)}$\\
        \bottomrule
        \end{tabular}
        }
        \caption{Metric 1: Success rate ($\%$). The mean score over $5$ runs is provided with the standard deviation as sub-scripts.}
        \label{tab: sr}
    \end{minipage}
    \begin{minipage}{.49\linewidth}
        \flushright
        \scalebox{0.52}{
        \begin{tabular}{lcc|ccccc|cc}\toprule
        & \multicolumn{4}{c}{$L\geq 1$} && \multicolumn{4}{c}{$L\geq 5$} \\
        \cline{2-5} \cline{7-10}
        \multirow{2}{*}{Models} & \multirow{2}{*}{$r_{bin}$} & $r_{base}$ & \multicolumn{2}{c}{ours} && \multirow{2}{*}{$r_{bin}$} & $r_{base}$ & \multicolumn{2}{c}{ours} \\
        \cline{4-5} \cline{9-10}
        & & \cite{pmlr-v155-pal21a} & $r_{depth}$ & $r_{bbox}$ && & \cite{pmlr-v155-pal21a} & $r_{depth}$ & $r_{area}$\\
        \hline
        GCN \cite{yang2018visual} & $10.0_{(0.4)}$ & $\mathbf{10.8_{(0.5)}}$ & $5.5_{(0.2)}$ & $8.2_{(0.1)}$ & & $10.3_{(0.7)}$ & $\mathbf{11.2_{0.7}}$ & $7.3_{(0.3)}$ & $8.7_{(0.3)}$\\
        SAVN \cite{Wortsman_2019_CVPR} & $11.0_{(0.2)}$ & $\mathbf{11.1_{(0.3)}}$ & $6.6_{(0.3)}$ & $10.5_{0.2}$ & & $11.7_{(0.1)}$ & $12.4_{(0.5)}$ & $10.5_{0.3}$ & $\mathbf{12.8_{(0.6)}}$\\
        M\_O \cite{pmlr-v155-pal21a} & $18.5_{(0.3)}$ & $\mathbf{20.7_{(0.2)}}$ & $11.6_{(0.1)}$ & $15.8_{0.4}$ & & $17.8_{(0.3)}$ & $\mathbf{20.0_{(0.6)}}$ & $13.7_{(0.3)}$ & $17.3_{(0.5)}$\\
        M\_R \cite{pmlr-v155-pal21a} & $24.4_{(0.3)}$ & $\mathbf{26.5_{0.2}}$ & $15.0_{(0.3)}$ & $16.8_{(0.2)}$ & & $26.2_{(0.4)}$ & $\mathbf{27.2_{(0.3)}}$ & $20.3_{(0.4)}$ & $19.3_{(0.4)}$\\
        \bottomrule
        \end{tabular}
        }
        \caption{Metric 2: SPL ($\%$). The mean score over $5$ runs is provided with the standard deviation as sub-scripts.}
        \label{tab: spl}
    \end{minipage}
\end{table*}

Reward shaping for reinforcement learning is a way to provide localized signals to an agent for encouraging behavior that is consistent with prior knowledge \cite{laud2004theory}. For the task of indoor robot navigation in search of a target object of interest, it is quite important for an agent to obtain intermediate auxiliary signals based on surrounding objects, to ensure that it's heading towards the goal. This is specially true for large environments, where the the robot may need to take a number of steps to reach the goal \cite{pmlr-v155-pal21a}. A popular reward function used in the object-goal navigation literature \cite{zhu2017target, yang2018visual, Wortsman_2019_CVPR, du2020learning, du2021vtnet} is of a binary nature, where a large positive reward is given at the goal state, while a smaller negative step penalty is assigned for every other state. Unfortunately, this type of a signal is quite sparse, thereby discouraging the learning process.

An alternate approach which has gained interest \cite{chaplot2020object, Maksymets_2021_ICCV} is to use geodesic distance to the closest target as a reward signal. Although this is a denser function compared to the binary reward, absolute knowledge about the closest distance to goal is a strong assumption that may not be easily available outside certain simulation environments \cite{Savva_2019_ICCV}. In contrast, we propose a method that relies on the estimated distance to objects calculated via different heuristics. Two approaches which are similar to ours are that of Druon \etal \cite{8963758} and Ye \etal \cite{ye2018active}. They both provide auxiliary signals based on the bounding box area of objects. However, these rewards are only assigned for the target object, and therefore, the signals are still quite sparse, specially when targets are smaller in size.

In this work, we build on the initial approach described in \cite{pmlr-v155-pal21a} by defining distance-based heuristics to modify the reward for both target objects, and other large, salient objects which have a close relationship with the target (called parent objects). In Section \ref{sec:method}, we describe two approaches for this. Next, in Section \ref{sec:expt_result}, we discuss the results obtained by utilizing the proposed reward shaping mechanism. Finally, we conclude with a discussion in Section \ref{sec:conclusion}.
\section{Methodology}
\label{sec:method}

Pal \etal \cite{pmlr-v155-pal21a} introduced a reward shaping mechanism where the agent receives a "partial" reward, $R_p$, when it can identify a parent object with close relationship to the target. This is given by ${R_p = R_t*Pr(t|p)*k}$, where $R_t$ is the target reward, and $Pr(t|p)$ is a probability distribution of the relative "closeness" of all the parent objects, $p$, to a given target object, $t$. Additional details can be found in \cite{pmlr-v155-pal21a}. Additionally, the scaling factor, $k$, is a constant kept fixed at $0.1$. Therefore, the partial reward is independent of the distance between the agent and the parent/target objects, $d$. Moreover, $R_p$ was only provided when the agent is within a distance threshold from the parent (set as $1$m in \cite{pmlr-v155-pal21a}). In this work, we propose \textit{two} methods to address these problems by reformulating $k$ as a factor of $d$. Furthermore, we extend the $R_p$ formulation towards both parent, and target objects. The primary motivations for this are: (i) the agent should be encouraged to identify parent objects whenever they are visible, and (ii) by making the reward a factor of $d$, the agent is further inspired to explore regions closer to $p$. 

\begin{figure}[h]
      \centering
      \begin{subfigure}[b]{0.234\textwidth}
          \centering
          \includegraphics[width=\textwidth]{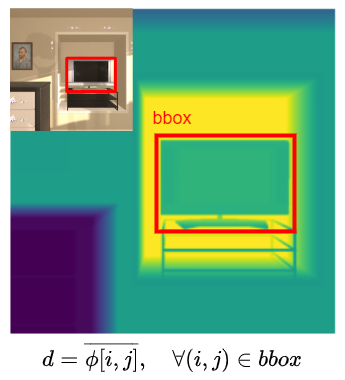}
          \caption{Metric distance from depth maps}
          \label{fig:depth_reward}
      \end{subfigure}
      \hfill
      \begin{subfigure}[b]{0.234\textwidth}
          \centering
          \includegraphics[width=\textwidth]{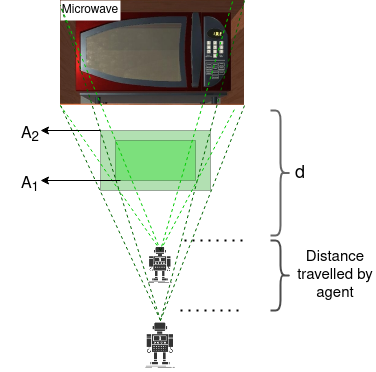}
          \caption{Relative distance from bbox area}
          \label{fig:bbox_reward}
      \end{subfigure}
      \hfill
         \caption{Image on the left shows depth map with a bounding box around the object. Inset contains the RGB image of the object. $d$ is obtained by finding the average distance of each pixel in the bounding box. Image on the right shows the relative increase in bounding box area of an object ($A_1$ to $A_2$) as the agent moves closer. $d$ is object distance when area is $A_2$.}
         \label{fig:rewards}
 \end{figure}

\textbf{(i) Utilizing metric depth} - Our first approach involves using metric depth in the form of depth maps obtained directly from the AI2-THOR simulator \cite{ai2thor_2017}. In lieu of this, an RGB-D sensor can also be used to get the estimated depth. From the depth maps, we compute $d$ as the average value of the region, $\phi$, bounded by an object's bounding box. This is illustrated in Figure \ref{fig:depth_reward}. Subsequently, the scaling factor is formulated as a linear function, $k'(d) = k*(m*d+c)$. In our experiments, $m=-0.15$, and $c=1$ were empirically chosen to ensure $k' \in [0,1]$.

\vspace{-0.15in}

\vspace{-0.15in}
\textbf{(ii) Utilizing bounding box area} - While the metric depth approach is intuitive, in theory, we observed that due to the added sensor input in the form of depth maps, the training time increased. Thus, our next approach was to use a heuristic for relative distance, where the scaling factor is calculated based on the assumption that as the agent moves closer, an object's bounding box (bbox) area should proportionately increase. This method, apart from being simple to implement, also reduces the dependence on additional sensor data, thereby minimizing the computational load. For this strategy, the scaling factor is given by $k'(d) = k*(1- (A_1/A_2 (d))^{0.5})$, where $A_1$ and $A_2$ are bounding box areas of a particular object in the state, when it was first seen by the agent and the current state respectively. This is depicted in Figure \ref{fig:bbox_reward}. 

In the next section, we validate our proposed hypothesis via extensive experiments.

\section{Experiments and Results}
\label{sec:expt_result}

We use the AI2-THOR \cite{ai2thor_2017} environment for our experiments. The setup and train/test split are consistent with other standard methods - GCN \cite{yang2018visual}, SAVN \cite{Wortsman_2019_CVPR}, and MJOLNIR-O/R \cite{pmlr-v155-pal21a}. We trained the agents for $3 \times 10^6$ episodes for each model. Furthermore, for every model, we conduct experiments using $4$ different reward functions - binary reward $r_{bin}$, baseline partial reward from \cite{pmlr-v155-pal21a}, $r_{base}$, and our two proposed rewards, namely depth-based, $r_{depth}$, and area-based, $r_{area}$, respectively. The evaluation metrics adopted from Anderson \etal \cite{evaluation_metric_2018}.

\textbf{Metric 1 discussion: Success rate (SR)} - Table \ref{tab: sr} shows the performance for this metric. For nearly every model, training via the proposed reward mechanism yields the best results, specially for episodes with larger path lengths, \ie $L \geq 5$, where further exploration of the environment might be needed. This shows the benefits of adding a denser reward signal based on distance to objects.

\textbf{Metric 2 discussion: Success weighted by Path Length (SPL)} - As opposed to the results for success rate, the SPL performance drops for the proposed methods. This is shown in Table \ref{tab: spl}. A possible reason for this could be due to the added incentive that the agent now gets to explore regions around parent objects, before heading towards the target. However, we do not necessarily view this as a major drawback, as exploring the environment is an important feature, specially in large and previously unseen environments.

It should also be noted that generally, the denser distance-based reward functions perform better for models that consider object relationships (like GCN \cite{yang2018visual}, and the MJOLNIRs \cite{pmlr-v155-pal21a}). This supports our intuition that adding auxiliary signal based on surrounding objects can aid in the search of far-off target objects.


\section{Conclusion}
\label{sec:conclusion}
We introduced a distance-based reward shaping mechanism that provides a denser feedback to the agent, thereby encouraging it to explore more of the environment. We showed that adopting this strategy leads to higher success rate of reaching the target object for multiple models, specially for cases where the optimal path requires taking a longer sequence of actions. However, due to the added exploration, the path length increases as a result. As part of our future work, we plan to address this issue by adopting imitation learning techniques \cite{du2020learning, du2021vtnet}.

{\small
\bibliographystyle{ieee_fullname}
\bibliography{egbib}
}

\end{document}